\documentclass[10pt,twocolumn,letterpaper]{article}
\date{}
\usepackage{booktabs}
\usepackage{graphicx}
\usepackage{amssymb}
\usepackage{amsmath}
\usepackage{url}
\usepackage[compact]{titlesec}
\usepackage{authblk}
\usepackage{balance}
\usepackage{color}
\usepackage[colorlinks=true, linkcolor=blue]{hyperref}
\usepackage{enumitem}
\usepackage{algorithm}
\usepackage[noend]{algpseudocode}

\usepackage[letterpaper,margin=0.75in]{geometry}
\title{Foundation Models for Zero-Shot Segmentation of Scientific Images without AI-Ready Data}

\author[1]{Shubhabrata Mukherjee}
\author[2]{Jack Lang}
\author[2]{Obeen Kwon}
\author[4]{Valerie Brogden}
\author[1,3]{Adam Weber}
\author[1,2]{Iryna Zenyuk}
\author[1,3]{Daniela Ushizima}
\affil[1]{Lawrence Berkeley National Laboratory, Berkeley, CA, USA}
\affil[2]{University of California, Irvine, USA}
\affil[3]{University of California, Berkeley, USA}
\affil[4]{Covalent Metrology, Sunnyvale, USA}

\begin{document}
\maketitle

\section{Abstract}
Zero-shot and prompt-based models have excelled at visual reasoning tasks by leveraging large-scale natural image corpora, but they often fail on sparse and domain-specific scientific image data. We introduce Zenesis, a no-code interactive computer vision platform designed to reduce data readiness bottlenecks in scientific imaging workflows. Zenesis integrates lightweight multimodal adaptation for zero-shot inference on raw scientific data, human-in-the-loop refinement, and heuristic-based temporal enhancement. We validate our approach on Focused Ion Beam Scanning Electron Microscopy (FIB-SEM) datasets of catalyst-loaded membranes. Zenesis outperforms baselines, achieving an average accuracy of 0.947, Intersection over Union (IoU) of 0.858, and Dice score of 0.923 on amorphous catalyst samples; and 0.987 accuracy, 0.857 IoU, and 0.923 Dice on crystalline samples. These results represent a significant performance gain over conventional methods such as Otsu thresholding and standalone models like the Segment Anything Model (SAM). Zenesis enables effective image segmentation in domains where annotated datasets are limited, offering a scalable solution for scientific discovery.
\footnote{This paper has been accepted for presentation at the 59th International Conference on Parallel Processing (ICPP 2025), DRAI workshop.}

\section{Introduction}

The rapid development of multimodal foundation models has reshaped computer vision, enabling zero-shot and prompt-based reasoning across diverse visual tasks. Yet, scientific domains remain underserved due to a persistent bottleneck: the inaccessibility of these models to raw scientific data. Most computer vision pipelines assume AI-ready inputs, i.e., datasets that are carefully curated for machine learning applications, with high-quality images, consistent formatting, dense and accurate annotations (e.g., labels, masks, or bounding boxes), and sufficient variability to support generalization. In contrast, scientific data are often sparse, heterogeneous, and lack standardized preprocessing, making it difficult to leverage the full potential of foundation models.

This data readiness challenge is particularly acute in materials science imaging, where instruments such as FIB-SEM, cryo-TEM~\cite{fornaciari2025achieving}, and micro-CT~\cite{Xu:2023} generate raw data with extreme bit depths, anisotropic voxel sizes, and domain-specific artifacts that fundamentally differ from the RGB natural scenes used to train contemporary foundation models \cite{ge2020deep}. Although these models excel in web-sourced imagery, they exhibit severe performance degradation when faced with the heterogeneous characteristics of scientific data, creating a significant barrier for domain experts seeking to leverage AI advances in their research workflows \cite{mazurowski2023segment}.

Traditional approaches require extensive preprocessing pipelines, format conversions, and domain-specific fine-tuning tasks that require substantial time, technical expertise, and manual processing. Existing platforms like ImageJ \cite{schneider2012nih} and CellProfiler \cite{carpenter2006cellprofiler} provide flexibility but lack integration with modern AI capabilities.

\begin{figure}[t]
\centerline{\includegraphics[width=0.4\textwidth]{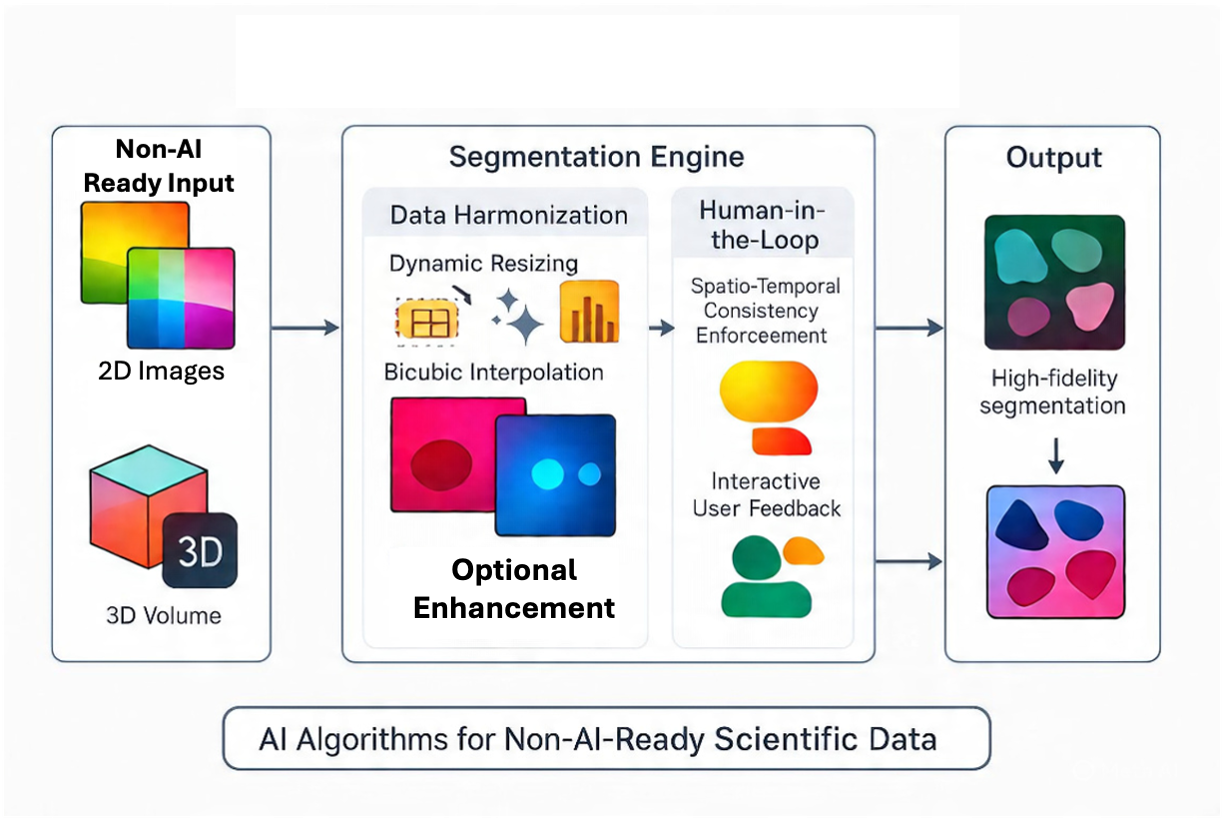}}
\caption{Transforming non-AI-ready scientific data}
\label{block}
\end{figure}

To address these fundamental challenges and democratize AI-powered visual reasoning for scientific research, we propose \textbf{Zenesis}, a comprehensive interactive no-code platform designed to enhance the data readiness of scientific images for AI applications. Our system seamlessly connects raw scientific data to AI-ready formats (Fig. \ref{block}), empowering domain experts that may use different image acquisition instruments to perform advanced segmentation tasks without needing extensive computer vision knowledge or specialized model training.

Zenesis provides native support for a wide range of scientific imaging modalities, including grayscale and RGB formats, 8/16/32-bit depth images, and both 2D and volumetric data. The platform integrates zero-shot segmentation capabilities for fully automated processing as well as with human-in-the-loop refinement options, such as interactive bounding box guidance and hierarchical segmentation for multi-level analysis. Additionally, it features a comprehensive evaluation dashboard that provides real-time quantitative assessments using standard metrics (DICE, IoU, accuracy) at both individual sample and dataset granularities.

We demonstrate the advantages of our approach through comprehensive validation on FIB-SEM material images, a particularly demanding use-case since the noisy raw data typically demands extensive preprocessing to achieve AI compatibility. This work exemplifies the broader objective of improving data readiness across scientific domains by making previously non-AI-compatible data usable through our method, which lowers the entry barrier; materials imaging serves as a representative case for advancing interactive visual reasoning. As segmentation is often the entry point in scientific image analysis workflows, Zenesis provides a reliable foundation for downstream tasks such as quantitative analysis of features like thickness, porosity, tortuosity, grain size distribution, and interface morphology.

\textbf{Our contributions are:}
\begin{enumerate}
    \item We formalize the challenge of enhancing \textbf{data readiness for non-AI-ready scientific images}, identifying key format, dimensional, and semantic incompatibilities with existing foundation models.
    \item We introduce \textbf{Zenesis}, a unified no-code platform that enables zero-shot and interactive segmentation across diverse scientific domains while preserving data fidelity.
    \item We develop lightweight \textbf{multi-modal adaptation techniques} that enable foundation models to operate accurately on raw scientific data without requiring extensive retraining.
    \item We integrate a comprehensive \textbf{real-time evaluation framework}, supporting quantitative evaluation at multiple granularities for robust validation of segmentation performance.
    \item We validate our system on challenging material science datasets, particularly FIB-SEM volumetric imagery, demonstrating practical operation on highly non-AI-ready scientific data.
\end{enumerate}

\section{Related Work}
SAM ~\cite{kirillov2023segment} and DINO(self-DIstillation with NO labels)~\cite{caron2021emerging}, and their combined forms like GroundingDINO~\cite{liu2024grounding}, represent significant strides in zero-shot computer vision technologies.
However, these advances remain largely inapplicable to scientific domains due to fundamental training biases toward natural RGB imagery from web sources. The resulting domain gap becomes particularly pronounced for scientific imaging modalities that exhibit non-standard characteristics such as high bit depths, specialized acquisition parameters, and domain-specific artifacts \cite{srinivas2024parameter}.

Recent research has highlighted the critical importance of data readiness as a prerequisite for effective AI deployment \cite{hiniduma2025data}. While data quality and preprocessing have received attention in general AI contexts, scientific imaging presents unique challenges that extend beyond conventional data preparation workflows. The heterogeneity of scientific imaging demands specialized approaches that preserve domain-specific information while enabling AI compatibility.

Interactive segmentation systems have emerged as a promising approach to address the brittleness of fully automated methods \cite{wang2018deepigeos}. However, existing interactive frameworks remain predominantly focused on medical imaging applications and rely heavily on domain-specific fine-tuning strategies \cite{wang2018interactive, zhu2024meduhip}. Despite sharing characteristics with other scientific domains, materials science data require specialized routines to handle extreme dynamic ranges, anisotropic sampling, and low signal-to-noise ratio.

Several specialized platforms have attempted to address scientific image analysis, with tools like IMOD ~\cite{kremer1996computer} for electron microscopy and domain-specific packages for astronomical data processing. However, these solutions demand extensive domain expertise for effective use. The integration of foundation model capabilities into scientific workflows remains an open challenge, and existing approaches often require substantial technical infrastructure and computational resources.

Recent advances in volumetric analysis have underscored the difficulty of extending 2D foundation models to 3D scientific data \cite{he2025vista3d, cciccek20163d}. Applying 2D models slice-by-slice often fails to preserve volumetric coherence, while fully 3D approaches require extensive computational resources and annotated data. These challenges are especially pronounced in materials science, where analyzing complex multi-phase microstructures in low-contrast 3D imagery is critical.

\section{Foundation Model for Segmentation}

SAM~\cite{kirillov2023segment} and SAM 2~\cite{ravi2024sam2} have revolutionized computer vision segmentation, enabling users to delineate objects using intuitive prompts such as point clicks, bounding boxes, and rough masks. These models leverage large-scale pre-trained models to adapt to diverse tasks with minimal user input.

SAM consists of an image encoder (Vision Transformer), prompt encoder, and mask decoder. SAM 2 extends this to video sequences with streaming memory mechanisms for real-time processing and temporal consistency. Examples of prompting mechanisms include: (a) Point-based prompting: users click on objects;
(b) Bounding box prompts: spatial constraints for object localization;
(c) Rough mask prompts: coarse segmentation outlines;
(d) Text-based prompting: CLIPSeg~\cite{luddecke2022image} and Grounding DINO + SAM~\cite{liu2024grounding,ren2024grounded} enable open-vocabulary segmentation and natural language-driven workflows.

In addition, there are specialized variants of SAM, such as:
a) FastSAM~\cite{zhao2023fast}: a YOLOv8-based architecture designed for efficiency;
(b) MobileSAM~\cite{zhang2023faster}: a lightweight alternative optimized for mobile deployment;
(c) MedSAM~\cite{ma2024segment}: an adaptation specifically for medical imaging;
(d) $\mu$SAM~\cite{archit2025segment}: an adaptation tailored for microscopy imaging.

Key challenges persist in computational efficiency, fine-grained segmentation, and domain adaptation. This work demonstrates how the zero-shot capabilities of multimodal foundation models, coupled with an interactive interface, can help address these issues, as detailed in the following section.

\begin{figure}[hbt!]
\centerline{\includegraphics[width=0.4\textwidth]{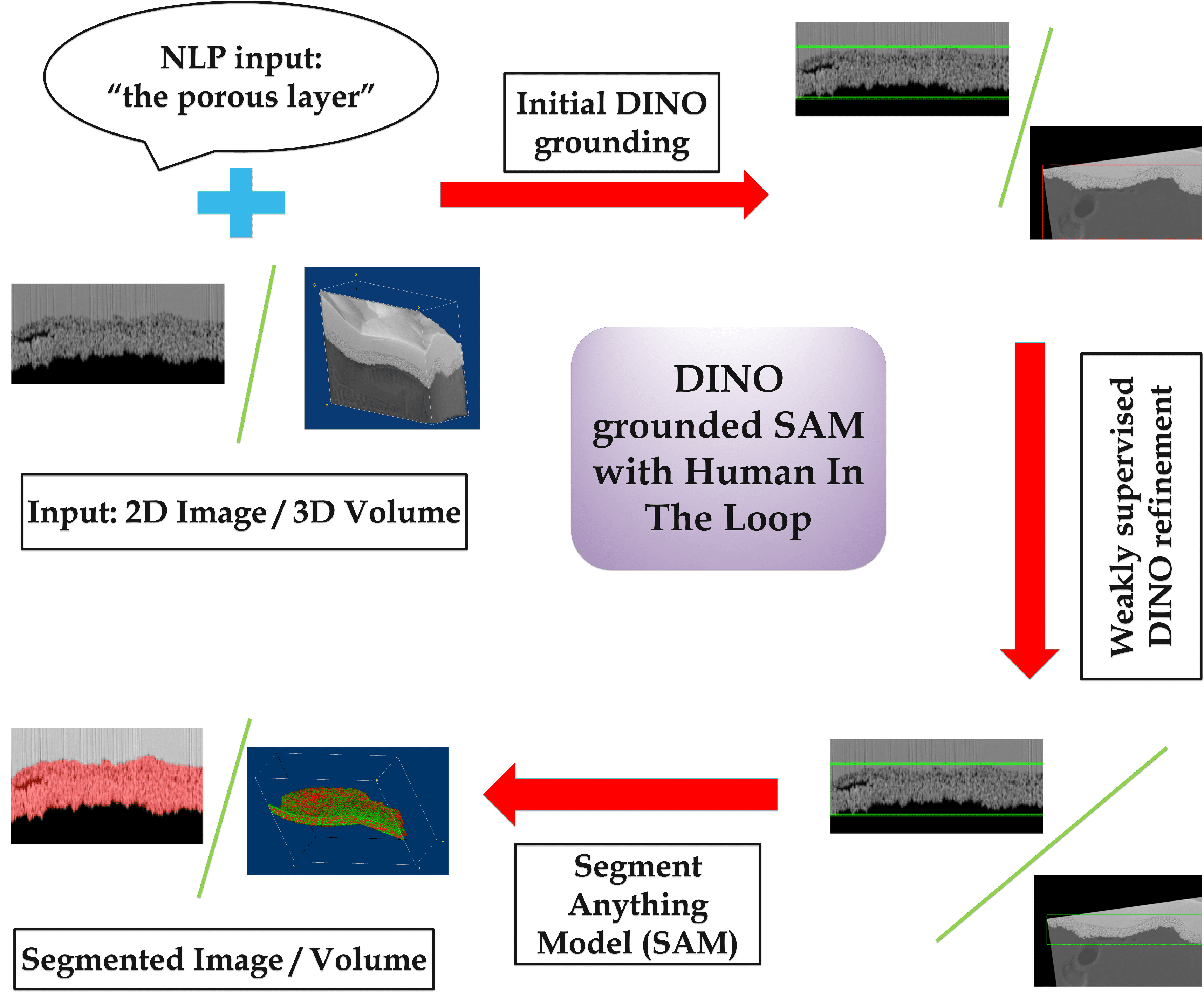}}
\caption{Interactive segmentation with DINO-SAM}
\label{dinosam}
\end{figure}

\subsection{Zenesis: DINO Grounded SAM with Human In-The-Loop}

Zenesis is designed as a modular, scalable web application enabling intuitive, text-driven image segmentation.

\begin{figure}[hbt!]
\centerline{\includegraphics[width=0.4\textwidth]{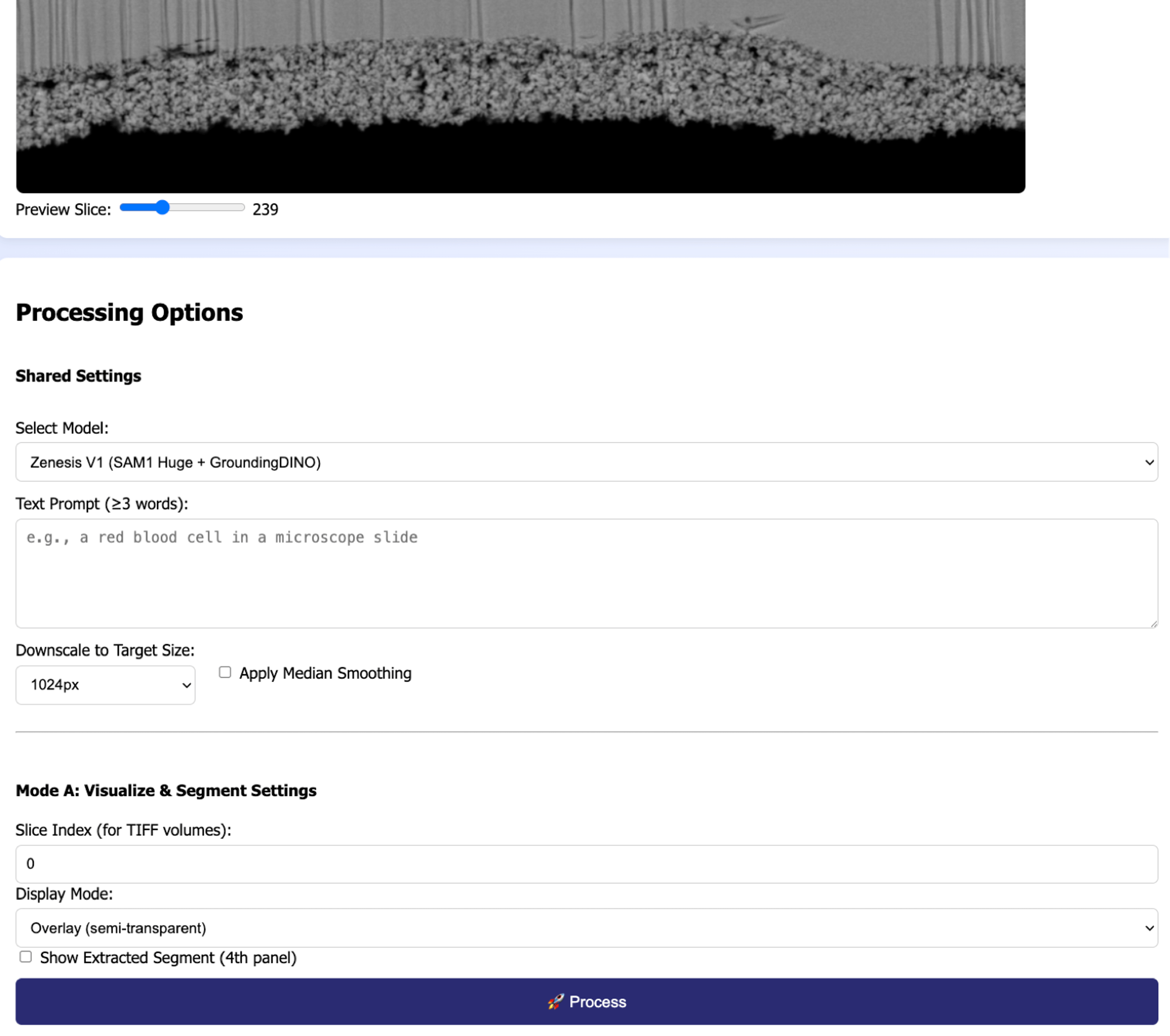}}
\caption{Interactive segmentation with Zenesis:\\
raw image / volume preview,natural language \\
prompt for segmentation}
\label{start}
\end{figure}

Zenesis's Presentation Layer (Fig.~\ref{start}) offers an interactive interface with three primary modes: Mode A for interactive segmentation of a single image or user-selected slice of a volume, Mode B for batch processing of volumes or multiple images, and Mode C for evaluation (Fig.~\ref{dashboard}). Through HTML templates and JavaScript controls, it dynamically renders input previews, visualizes DINO bounding boxes and segmentation mask overlays (or highlighted segment boundaries), and optionally displays extracted segments. In practice, generating a segmented result requires just three steps: (1) uploading an image or volume, (2) entering a natural language prompt, and (3) clicking ``Process'' to view the segmented overlay that is typically completed in under a few seconds with default settings.

The Core Processing Pipeline integrates a transformer-based DINO model for text-conditioned bounding-box generation and a SAM model for refining boxes into precise segmentation masks. For our implementation, we employed the GroundingDINO model with a Swin-T variant and the Segment Anything Model (SAM) with a ViT-H backbone. It also features an optional sliding-window average-based thresholding for outlier correction in volumetric data, enhancing temporal and spatial consistency across slices.

\begin{figure*}[hbt!]
\centerline{\includegraphics[width=\linewidth]{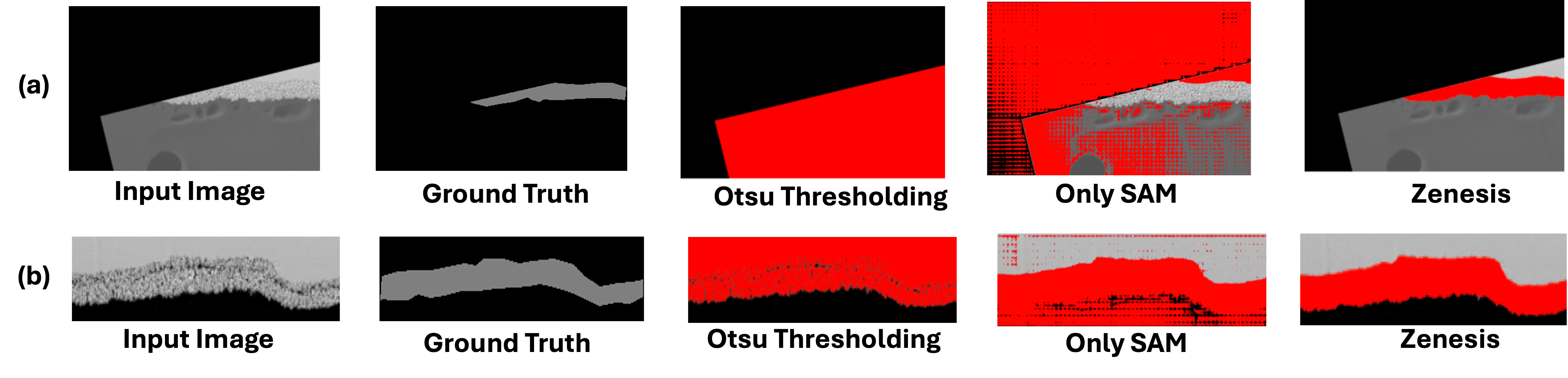}}
\caption{Catalyst Layer Segmentation by Otsu, SAM, and Zenesis: (a) Crystalline, (b) Amorphous}
\label{comparison}
\end{figure*}

\subsection{Theoretical Framework and Workflow}

\begin{figure*}[hbt!]
\centerline{\includegraphics[width=0.9\textwidth]{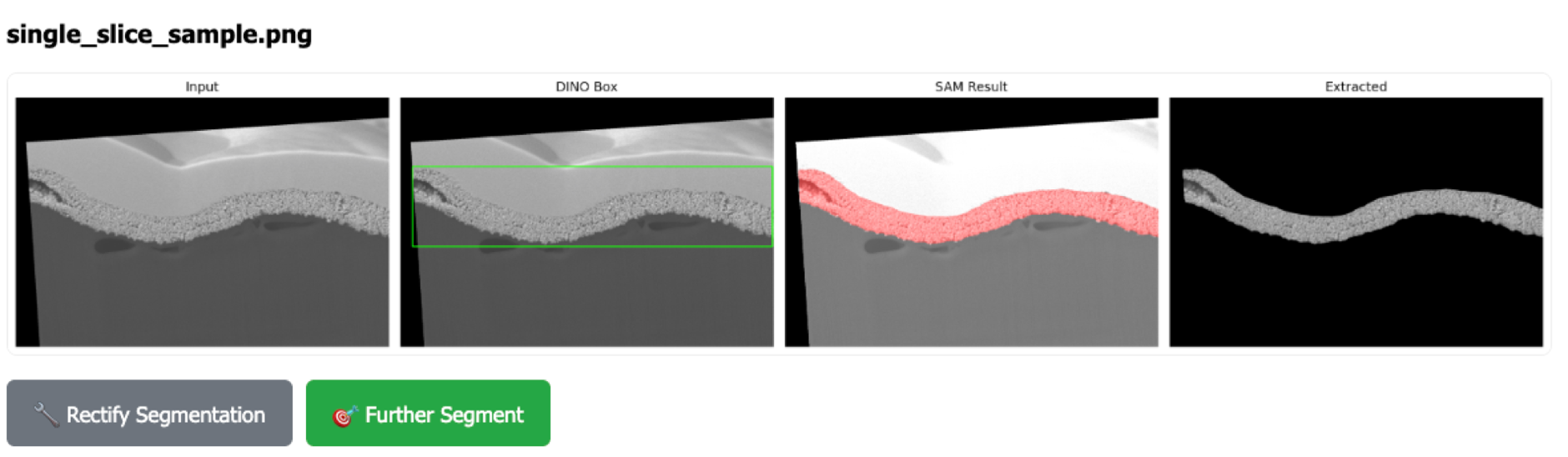}}
\caption{DINO bounding box prediction, segmentation overlay, and segmented catalyst layer; bottom buttons indicate option for \textbf{Rectify Segmentation} and \textbf{Further Segment}.}
\label{segm_example}
\end{figure*}


\begin{figure}[hbt!]
\centerline{\includegraphics[width=\linewidth]{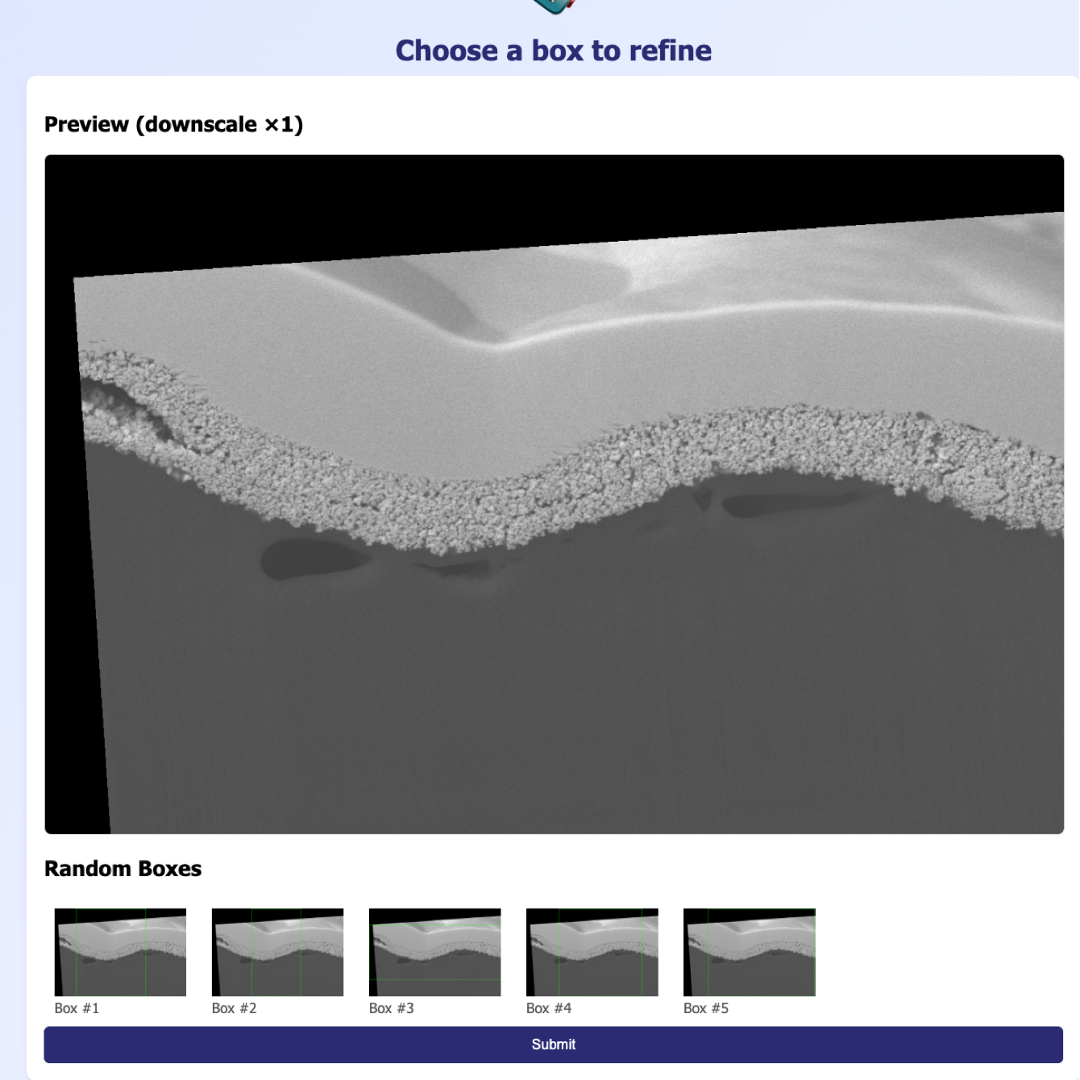}}
\caption{DINO Box Refinement with random boxes allows interactive user feedback.}
\label{boxes}
\end{figure}

This section details the underlying theory and the step-by-step operation of the Zenesis pipeline Fig.~\ref{dinosam}, highlighting GroundingDINO, SAM, human-in-the-loop corrections and volumetric refinement.

Zenesis employs a transformer-based GroundingDINO encoder to project text prompts and image inputs into a shared embedding space. Cross-modal attention then computes relevance scores between text tokens (queries) and image patch embeddings (keys and values) as follows:

\[
\mathrm{Attention}(Q,K,V) = \mathrm{softmax}\left(\frac{QK^T}{\sqrt{d}}\right)V
\]

where Q, K, and V represent query, key, and value matrices derived from the text and image embeddings. High-confidence regions are output as bounding boxes, controlled by box and text thresholds, allowing natural language guidance of object localization.

\begin{figure*}[hbt!]
\centerline{\includegraphics[width=\linewidth]{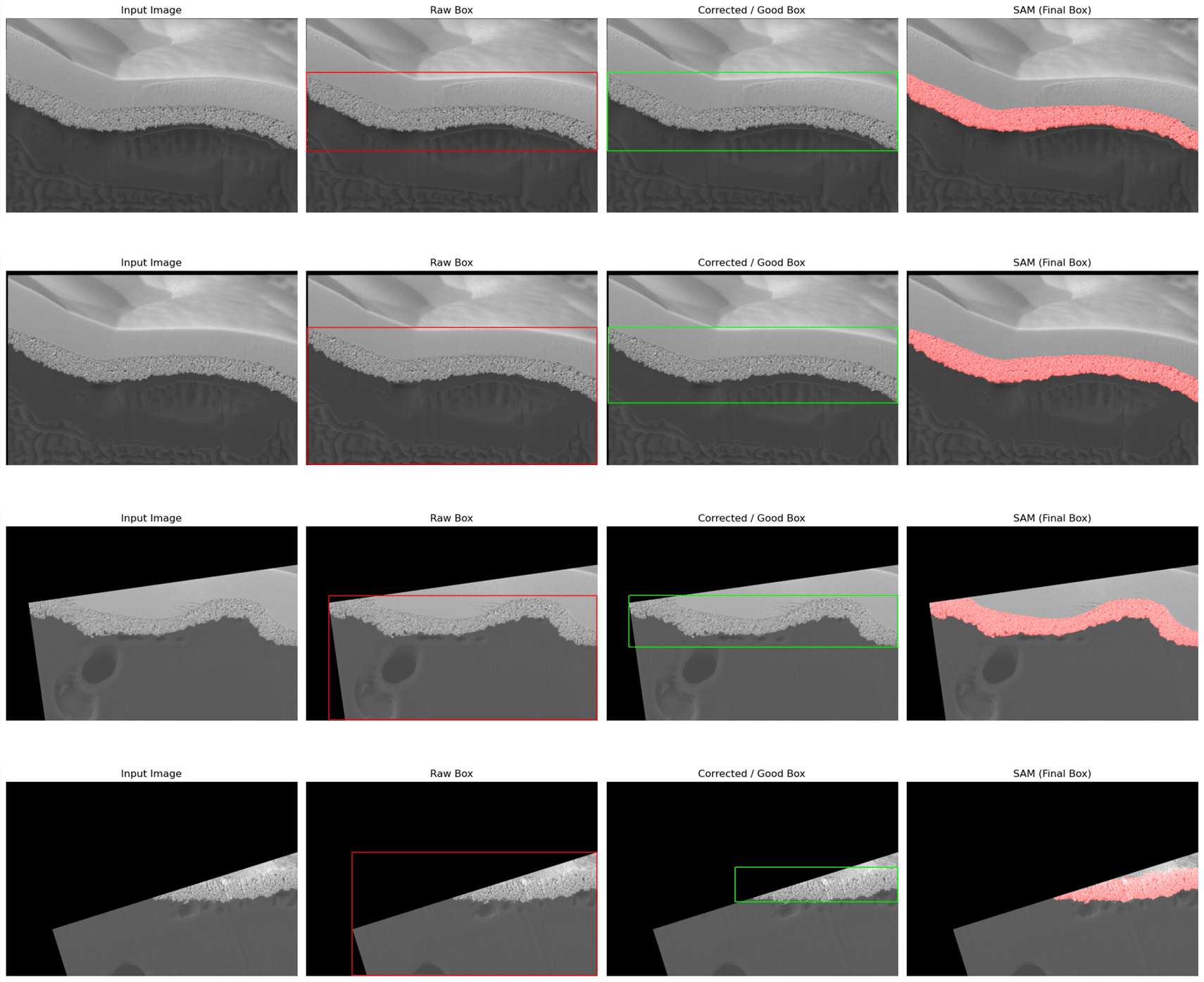}}
\caption{DINO Box Refinement process for volumes with heuristic thresholding}
\label{correct}
\end{figure*}

SAM receives GroundingDINO-derived bounding boxes as input and applies its segmentation head to generate fine-grained masks. By considering multi-scale prompts (boxes), SAM produces stable, pixel-level segmentation, which are overlaid onto the original image for user interpretation. To address bounding-box outliers, especially in volumetric sequences, Zenesis implements an interactive correction mechanism:
\begin{enumerate}[nosep]
    \item Rectify Segmentation: \textbf{Human-in-the-loop} based adjustment of bounding boxes(Fig.~\ref{boxes}) allows users to generate random boxes (with criteria such as length or width equal to the image size) and select the nearest segmentation area of interest, providing a weakly supervised way to correct automated detections.
    \item Automated Heuristic Refinement: For multi-slice volumes, the system computes mean width/height across a fallback window of adjacent slices. Boxes exceeding a height or width factor are replaced (Fig.~\ref{correct})
    by the average box of previous slices, ensuring temporal consistency and mitigating artifacts due to sudden changes in appearance or GroundingDINO failures.
\end{enumerate}

Additionally, Zenesis offers another feature \textbf{Further Segment (Fig.~\ref{segm_example})}, which enables users to further inspect selected segments, allowing for hierarchical segmentation by triggering GroundingDINO and SAM on subregions for more detailed analysis.

Zenesis is a web-based platform that supports natural language-guided segmentation and refinement, allowing users to perform advanced scientific image analysis without needing ML or code programming expertise. Its interactive features—including random box-based mask refinement, volume-aware heuristics, and prompt-driven segmentation are accessible through an intuitive UI, making it suitable for both domain experts and non-technical users. Additionally, its built-in evaluation mode allows rapid feedback from limited labeled data, streamlining integration into downstream AI pipelines.

\section{Dataset Description}

\begin{figure}[hbt!]
\centerline{\includegraphics[width=\linewidth]{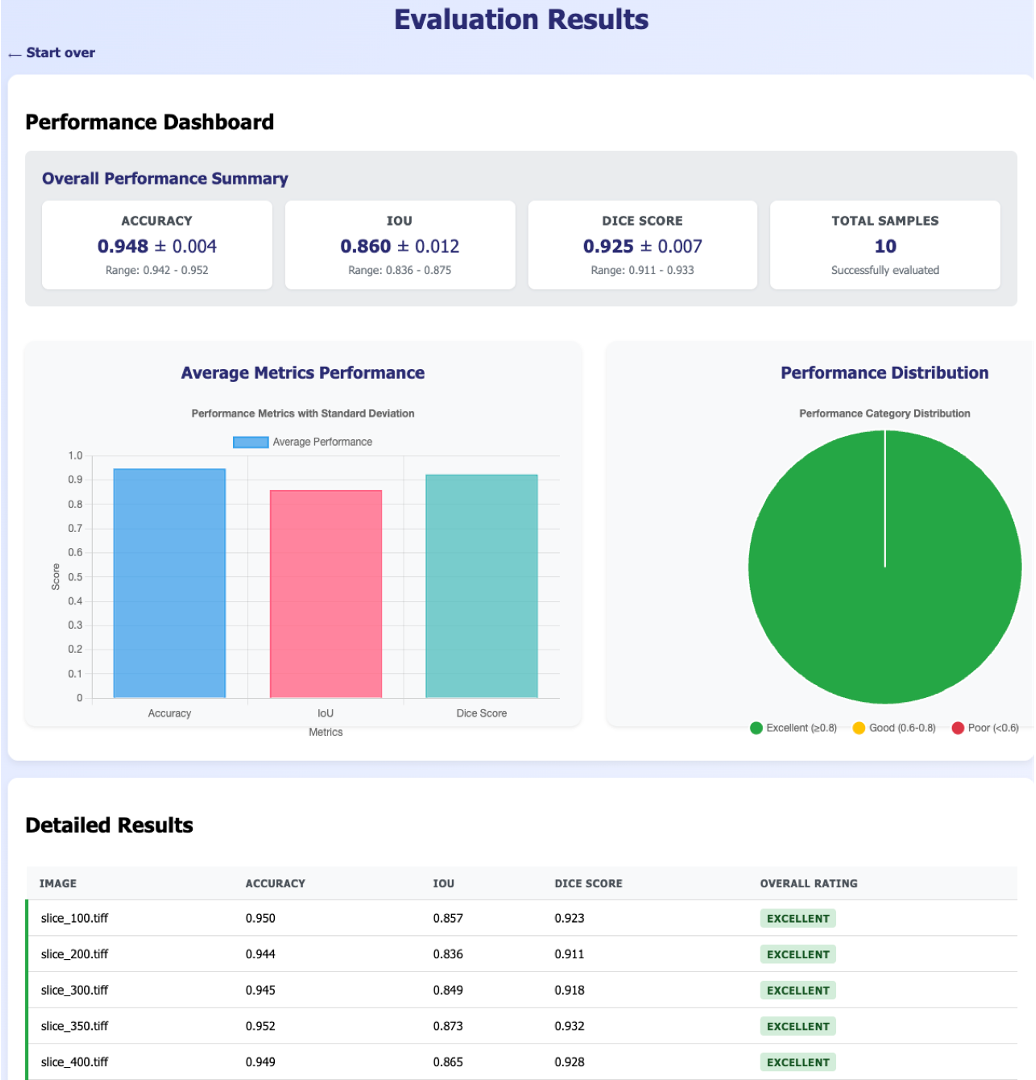}}
\caption{Segmentation performance dashboard}
\label{dashboard}
\end{figure}

The samples contain two types of iridium oxide catalysts: amorphous IrO$_2$ and crystalline IrO$_2$, embedded within thin films of Nafion D2021 ionomer. In the catalyst layer, the ionomer binds the catalyst particles and enables proton conduction, which is essential for the operation of Proton Exchange Membrane (PEM) electrolyzers.

This dataset was collected to improve understanding of catalyst loading and ionomer distribution in PEM catalyst layers~\cite{fornaciari2025achieving,kwon2024understanding}, which are critical for efficient water electrolysis. The dataset presents several challenges, including the lack of ground-truth annotations, high noise levels in grayscale images, hindering visual separation of material phases, and variability in contrast caused by defocus and sample topography.

Image acquisition was performed using Focused Ion Beam Scanning Electron Microscopy (FIB-SEM), a technique particularly suited for imaging materials that are sensitive to electron beams. The use of low-dose FIB-SEM allows for high-resolution imaging of thin-film ionomer morphology and coverage across the electrode structure, providing insights that are difficult to obtain with conventional electron microscopy techniques. These samples reveal distinctive structural properties, including needle-like morphologies in crystalline IrO$_2$ with a significantly higher specific surface area ($\sim$110 m$^2$ g$^{-1}$) compared to amorphous IrO$_x$ ($\sim$50 m$^2$ g$^{-1}$).

Electrochemical performance evaluations demonstrate important differences between the catalysts, with amorphous IrO$_2$ exhibiting distinct redox peaks indicative of surface activation processes, whereas crystalline IrO$_2$ primarily shows double-layer capacitance behavior at lower voltages. Both catalysts demonstrate competitive polarization performances, with voltages of 1.92 V (amorphous) and 1.94 V (crystalline) at 5 A cm$^{-2}$ for a loading of 0.85 mg cm$^{-2}$, accompanied by low Tafel slopes (43.8 mV dec$^{-1}$ for amorphous IrO$_x$, 53.7 mV dec$^{-1}$ for crystalline IrO$_2$).
For benchmarking purposes, this dataset utilizes 20 full slices extracted from 3D volumetric images, with 10 slices each from the crystalline and amorphous volumes, represented as 2D images derived from the original 3D TIFF files. This dataset significantly contributes to the broader mission of advancing electrochemical research by providing detailed structural and compositional information that aids in the design of next-generation energy devices with improved properties.

\section{Results and Analysis}

Zenesis, our proposed multimodal foundation model-based approach for interactive zero-shot segmentation, demonstrates excellent performance on both crystalline and amorphous datasets of FIB-SEM images. In comparison to Otsu thresholding and SAM-only approaches, Zenesis achieves superior segmentation performance. The average performance metrics for each dataset are presented in Table \ref{tab:combined_performance_metrics}.

\begin{table*}[h!]
    \centering
    \caption{Average Performance Metrics for Different Methods applied to Catalyst Layer}
    \label{tab:combined_performance_metrics}
    \tiny
    \resizebox{\textwidth}{!}{%
    \begin{tabular}{|l|ccc|ccc|}
    \hline
    \textbf{Method} & \multicolumn{3}{c|}{\textbf{Crystalline}} & \multicolumn{3}{c|}{\textbf{Amorphous}} \\
     & Accuracy & IOU & Dice & Accuracy & IOU & Dice \\
    \hline
    Otsu threshold & 0.586±0.125 & 0.161±0.057 & 0.274±0.080 & 0.581±0.019 & 0.407±0.024 & 0.578±0.024 \\
    SAM-only        & 0.485±0.146 & 0.100±0.083 & 0.173±0.137 & 0.499±0.160 & 0.405±0.088 & 0.571±0.087 \\
    \hline
    \textbf{Zenesis}         & \textbf{0.987±0.005} & \textbf{0.857±0.029} & \textbf{0.923±0.017} & \textbf{0.947±0.005} & \textbf{0.858±0.015} & \textbf{0.923±0.009} \\
    \hline
    \end{tabular}%
    }
\end{table*}

We evaluated segmentation performance using standard metrics including Intersection-over-Union (IoU) ~\cite{everingham2010pascal} and Dice coefficient ~\cite{dice1945measures}. IoU measures the overlap between predicted and ground truth regions as the ratio of intersection to union, while the Dice coefficient quantifies similarity as twice the intersection divided by the sum of the areas. The results demonstrate that Zenesis achieves significantly higher accuracy, IOU, and Dice scores for both crystalline and amorphous datasets compared to Otsu thresholding and SAM-only approaches. For the amorphous dataset, Zenesis attains an average accuracy of 0.947, an IOU of 0.858, and a Dice score of 0.923, while for the crystalline dataset, it achieves 0.987 accuracy, 0.857 IOU, and 0.923 Dice score. These results confirm Zenesis’s robust performance across both sample types.

In contrast, Otsu thresholding and SAM-only approaches struggle with the crystalline dataset. Otsu thresholding yields very low IOU (0.161) and Dice scores (0.274) for crystalline samples, while SAM-only fails entirely, producing 0.100 IOU and 0.173 Dice scores. This failure arises because both methods default to following the sharp gradient change in entirely black background as the region of interest due to the lack of distinct edges or intensity variations in crystalline structures. For the amorphous dataset, SAM-only performs better (accuracy: 0.499, IOU: 0.405, Dice: 0.571) but still lags behind Zenesis. Zenesis overcomes these limitations by leveraging GroundingDINO-based text-guided grounding, which provides precise bounding boxes to guide SAM’s segmentation. This integration ensures SAM focuses on the actual objects of interest rather than defaulting to background regions. For crystalline samples, this guidance enables Zenesis to achieve 86\% IOU and 92\% Dice scores, resolving the critical failure modes of Otsu and SAM-only.

The performance disparity between datasets likely stems from inherent differences in image characteristics. The amorphous dataset’s more distinct features (e.g., contrast, texture) allow models like SAM-only to perform better, while the crystalline dataset’s uniform and complex structures require the explicit guidance provided by Zenesis. The use of GroundingDINO in Zenesis is pivotal, as it addresses the core weakness of SAM and Otsu: their reliance on maximum confidence scores to select regions, which fails in (Fig. \ref{comparison}) low-contrast or ambiguous scenarios. By grounding segmentation to text-prompted objects, Zenesis reduces false positives and improves accuracy, particularly in challenging cases like crystalline samples.

Although this paper focuses on FIB-SEM datasets as a representative challenge in materials science, Zenesis is already being evaluated on other scientific imaging modalities such as cryo-TEM and micro-CT, with preliminary results indicating comparable or even greater segmentation improvements, highlighting the method's adaptability across domains with diverse image characteristics. The platform's built-in evaluation system (Fig.~\ref{dashboard}) supports performance validation using limited labeled data from any new scientific dataset, facilitating generalization assessment with minimal manual overhead.

A timing test on CPU-based systems shows consistent performance across varied image resolutions: a 640×480 RGB image from RefCOCO took ~8 seconds, a 292×1318 8-bit grayscale FIB-SEM image ~9 seconds, and a 2560×2560 32-bit micro-CT image 10 seconds, all using the SAM1 configuration on an Apple M4 Max CPU without GPU acceleration. These results suggest resolution-agnostic behavior under default settings and demonstrate interactive feasibility on consumer hardware. To enhance scalability, we are developing a GPU-accelerated SAM2-based version for deployment on the NERSC Perlmutter supercomputer, leveraging NVIDIA A100 GPUs. Prior benchmarks have shown 10–40X speedups for transformer-based vision models on similar architectures\cite{lakshminarasimhan2024brickdl}, indicating that Zenesis could achieve sub-second per-slice processing, enabling efficient segmentation of large-scale scientific datasets.

Overall, Zenesis shows exceptional adaptability and reliability, even in slices with complex or subtle features. This makes it a promising tool for real-world applications where high-quality annotations are scarce. Future work could explore optimizing the text-guided grounding process and extending the framework to other domains with similar segmentation challenges.

\section*{Conclusion and Future Research}

In this work, we introduced \textbf{Zenesis}, an interactive, no-code segmentation tool that enhances data readiness by transforming raw imaging inputs into semantically meaningful segments. It also accelerates the data readiness phase essential for training foundation models. By leveraging multimodal transformer-based techniques, including GroundingDINO and SAM, Zenesis enables accurate, zero-shot segmentation of complex scientific images such as FIB-SEM without requiring extensive preprocessing or model retraining. 

Our results demonstrate that Zenesis outperforms traditional methods like Otsu thresholding and SAM-only approaches, particularly in noisy, high bit-depth imaging domains. We validated Zenesis on challenging datasets featuring both crystalline and amorphous catalyst structures, achieving state-of-the-art performance in terms of accuracy, IoU, and Dice score. These results highlight Zenesis’s robustness and adaptability across diverse scientific imaging modalities, positioning it as a powerful tool for accelerating scientific discovery in domains traditionally under-served by AI models.

Looking ahead, several promising research directions emerge. First, we plan to extend Zenesis to support additional imaging modalities such as {X-ray diffraction (XRD)}, {scanning tunneling microscopy (STM)}, and {energy-dispersive X-ray spectroscopy (EDX)}, broadening its applicability across scientific domains. Second, we will develop support for {multi-object segmentation} within individual images and volumes, enabling more complex scene understanding. Third, we intend to introduce an optional {fine-tuning module} that allows advanced users to adapt the segmentation pipeline to highly specialized or critical datasets where domain-specific refinement is essential. To further demonstrate Zenesis's broader applicability and benchmark against established standards, we will evaluate a RefCOCO-style generalized referring expression-based segmentation benchmarking framework ~\cite{kazemzadeh2014referitgame}, which could enable large-scale validation using multimodal inputs (e.g., natural language + image) over expansive datasets, as established in vision-language segmentation literature. Through Zenesis, we take a step forward toward democratizing advanced AI capabilities for scientific imaging, facilitating broader access to interactive segmentation tools that preserve domain fidelity, reduce annotation burden, and accelerate hypothesis-driven research.

\section*{Acknowledgements}
This work was partially supported by the US Department of Energy (DOE) Office of Science Advanced Scientific Computing Research (ASCR), funding project Autonomous Solutions for Computational Research with Immersive Browsing \& Exploration (ASCRIBE). It also included partial support from DOE ASCR and Basic Energy Sciences (BES) under Contract No. DE-AC02-05CH11231 to the Center for Ionomer-based Water Electrolysis (CIWE) and the Center for Advanced Mathematics for Energy Research Applications (CAMERA). We also thank professor Shannon Boettcher and his team at UC Berkeley for their insightful discussions and technical assistance.

\balance
\bibliographystyle{plain}
\bibliography{bibliography}
\end{document}